\documentclass[runningheads]{llncs}
\usepackage{graphicx}
\begin{document}
\title{LexGPT 0.1: pre-trained GPT-J models with Pile of Law}
%\title{LegalGPT 0.1: pre-trained GPT-J models with legal text}

%generative pre-trained transformer for legal text}
\author{Jieh-Sheng Lee\orcidID{0000-0002-0990-6170}}
\authorrunning{J. Lee}
\institute{National Yang Ming Chiao Tung University School of Law \\
No. 1001, Daxue Rd. East Dist., Hsinchu City 300093, Taiwan
\\
\email{jasonlee@nycu.edu.tw}\\
}

\maketitle              % typeset the header of the contribution

\begin{abstract}
%The abstract should briefly summarize the contents of the paper in

This research aims to build generative language models specialized for the legal domain. The manuscript presents the development of LexGPT models based on GPT-J models and pre-trained with Pile of Law. The foundation model built in this manuscript is the initial step for the development of future applications in the legal domain, such as further training with reinforcement learning from human feedback. Another objective of this manuscript is to assist legal professionals in utilizing language models through the ``No Code'' approach. By fine-tuning models with specialized data and without modifying any source code, legal professionals can create custom language models for downstream tasks with minimum effort and technical knowledge. The downstream task in this manuscript is to turn a LexGPT model into a classifier, although the performance is notably lower than the state-of-the-art result. How to enhance downstream task performance without modifying the model or its source code is a research topic for future exploration.

\end{abstract}

\section{Introduction}
\label{section:introduction}

The codename ``LexGPT'' in this research refers to the development of GPT (Generative Pre-trained Transformer)~\cite{openai_gpt2_blog} models specialized for the legal domain. The objective is to create models that can assist legal professionals in performing various legal tasks in the future. In this manuscript, ``LexGPT 0.1'' refers to the models based on GPT-J~\cite{gpt-j-github} and pre-trained with Pile of Law~\cite{pile-of-law} dataset. LexGPT as a foundation model is essential for the development and success of future applications in the legal domain, including those might build on InstructGPT~\cite{instruct-gpt} or ChatGPT~\cite{openai_chatgpt_blog} models.
The progress made in this manuscript represents the initial step towards developing future models and applications. The pre-trained LexGPT models will be released for further research and development. 

To facilitate the adoption of the LexGPT models by legal professionals, a downstream classification task is presented as an example. This task involves fine-tuning the models without the need to add a classification layer, thereby eliminating the need for enhancing the source code of the model. 
In the legal domain, developing language models or applications may require technical skills that legal professionals may not possess. This can create a significant entry barrier for those who want to leverage LexGPT models in this research. 
To overcome this barrier, an objective of this manuscript is to explore the possibilities of leveraging language models under the ``No Code'' idea. In computer science, ``No Code'' refers to a way of building software applications without requiring extensive knowledge or experience in programming languages. The idea is intended to democratize access to technology by providing users with tools, templates, and interfaces to create applications easily and quickly. In this manuscript, the downstream tasks are conditioned on fine-tuning models with specialized data and without modifying any source code. By doing so, legal professionals can create custom language models based on pre-trained LexGPT models with minimum effort and technical knowledge.

\section{Related Work}
\label{section:related_work}

Language models have proven to be effective in many domains and are beginning to make an appearance in the field of law. For example, LEGAL-BERT~\cite{chalkidis-etal-2020-legal} shows that pre-training BERT from scratch with legal corpuses performs better on average. Without domain adaptation, the authors found that the previous pre-training and fine-tuning do not always generalize well in the legal domain. In~\cite{Darji_2023}, the authors fine-tuned a popular BERT language model trained on German data (German BERT) for Named Entity Recognition (NER) tasks in the legal domain.
In~\cite{pile-of-law}, a BERT-large equivalent model was trained to predict whether a paragraph should use pseudonymity. Legal datasets are scarce and expensive because of the complexity and specialty. The authors of~\cite{pile-of-law} curated a $\sim$256GB dataset of legal and administrative text, which is called Pile of Law. The dataset is intended for learning responsible data filtering from the law. In~\cite{lexglue}, the authors introduced the Legal General Language Understanding Evaluation (LexGLUE) benchmark, a collection of datasets for evaluating model performance across a diverse set of legal NLU tasks. The models evaluated in~\cite{lexglue} are BERT-based models. For processing long legal documents, the authors in~\cite{mamakas-etal-2022-processing} modified a Longformer warm-started from LegalBERT and modified LegalBERT to use TF-IDF representations. 

Most language models utilized in the legal field are built upon BERT model. While GPT models are adept at generative tasks, they are not commonly examined as a foundation model for legal tasks in academics. In~\cite{jiehsheng03}, the author fine-tuned OpenAI GPT-2~\cite{openai_gpt2_blog} models for patent claim generation. However, the models are specific to the patent field only. In~\cite{LawGPT_1.0}, the author built LawGPT 1.0 as a virtual legal assistant built by fine-tuning GPT-3 for the legal domain. The author provided a brief overview but the detailed information about the model is protected by a non-disclosure agreement (NDA) and cannot be disclosed. In~\cite{pile-of-law}, the authors noted that the Pile of Law dataset can be used in the future for pretraining legal-domain language models. Given the limited exploration of GPT models in the legal domain, this research is motivated to undertake pre-training of GPT models on the Pile of Law dataset for downstream legal tasks. The pre-training can serve as a precursor to the subsequent development of advanaced models or applications.
For evaluating the performance of natural language processing (NLP) models, the General Language Understanding Evaluation (GLUE)~\cite{glue} benchmark is a popular benchmark. In the legal domain, LexGLUE~\cite{lexglue} is a benchmark dataset for evaluating legal language understanding. Language models in~\cite{lexglue} rely on BERT only and no GPT models are included. For pre-training sizeable GPT models with open-sourced code, the repositories available in public include GPT-J-6B~\cite{gpt-j-github} (using TPUs), GPT-NeoX-20B~\cite{gpt-neox-20b} (using GPUs), and Open Pre-trained Transformers (OPT)~\cite{opt} (using GPUs).

\section{Implementation }

\subsection{Objectives}
\label{subsection:objectives}

The primary objective of this manuscript is to build LexGPT as foundation models by pre-training GPT models exclusively with legal text. It is important for legal professionals that the model generates only legal text. These pre-trained models will serve as the basis for downstream tasks. 
The second objective is to evaluate the performance of downstream tasks by fine-tuning the LexGPT models and using the LexGLUE benchmark. The tasks are developed under the purposeful constraint that no additional source code or new layers are added to the model.
Lastly, this manuscript aims to document instances of failure and lessons learned so that subsequent researchers may discover improved solutions. 

\subsection{Pre-trained models}
\label{subsection:pre-trained_models}

\subsubsection{Why GPT-J-6B?}
At the beginning of this research, the GPT-J-6B model in~\cite{gpt-j-github} was released as the largest pre-trained model available to the public. GPT-J-6B is a transformer model trained using Mesh Transformer JAX and developed by EleutherAI, an independent research organization focused on advancing open-source artificial intelligence. The model implements the GPT architecture developed by OpenAI. According to~\cite{gpt-j-github}, GPT-J-6B has achieved impressive results on various language tasks, such as text generation, translation, and question answering. The size of the GPT-J-6B model is also suitable for quicker iterations and proof of concept. The model runs on TPU (Tensor Processing Unit) instead of GPU.

\subsubsection{Why pre-training?}
The primary objective of this manuscript is to construct GPT models using the Pile of Law dataset. The codebase in~\cite{gpt-j-github} provides a guide for fine-tuning~\cite{gpt-j-how-to-fine-tune} the GPT-J-6B model. However, since the original GPT-J-6B model's training data does not solely include legal text, a fine-tuned model could generate non-legal text, which would be of little use to legal professionals. Hence, after referencing the fine-tuning guide, the models in this manuscript are pre-trained from scratch instead. These pre-trained models can serve as the foundation models for downstream tasks and future applications in the legal domain. For instance, an application such as ChatGPT requires a foundation model for training with reinforcement learning from human feedback (RLHF)~\cite{NIPS2017_d5e2c0ad}. The concept of incorporating human feedback to solve deep reinforcement learning tasks was introduced by OpenAI. The idea paved the way for developing InstructGPT, and now ChatGPT. 
To implement RLHF with legal professionals' feedback, the pre-trained LexGPT models serve as the first step towards this goal.

\subsubsection{Dataset}
In this manuscript, the Pile of Law dataset from~\cite{pile-of-law} is utilized to pre-train GPT-J models. The dataset was initially released with a size of 256G and is still expanding. As of the time of writing, the estimated size of the dataset is 291.5 GiB. The first release of the dataset is employed in this study.  The Dataset Card~\cite{dataset_card_pile_of_law} indicates that the dataset can be used for pre-training language models as a key direction in access-to-justice initiatives. 

\subsubsection{Tokenizer}
The GPT-J-6B model is trained with a tokenization vocabulary of 50257, using the same BPE (Byte Pair Encoding) as GPT-2 and GPT-3. To enhance the accuracy of language models, the LexGPT models discussed in this manuscript utilize domain-specific vocabularies trained from the Pile of Law. One of the tokenizers is trained with a vocabulary size of 50257, while the other has a reduced size of 25129, representing half of the former. These two tokenizers are provided for experiments.

\subsubsection{Model sizes \& Training Losses}
\label{subsection:model_size}

\begin{figure}[h]
  {\includegraphics[width=0.9\textwidth, keepaspectratio]{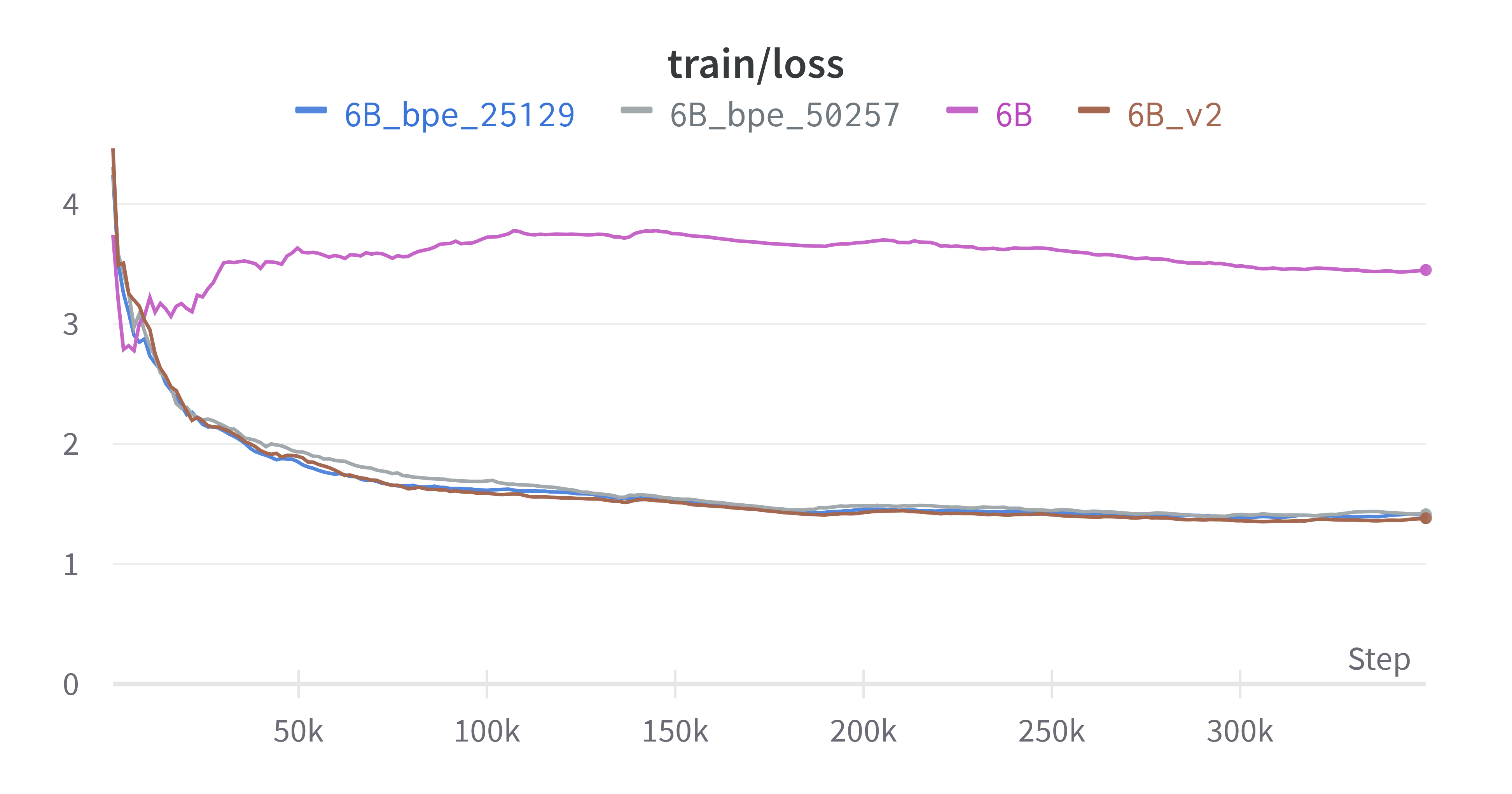}}
  %{\includegraphics[width=0.8\textwidth, keepaspectratio]{training_loss.png}}
  \caption{Training loss}
  \label{train_loss_1}
  %\label{training_loss}
\end{figure}

The training losses of LexGPT-6B models, using different tokenizers, are depicted in Fig.\ref{train_loss_1}. The model using the original tokenizer in GPT-J-6B is represented by the curve ``6B''. However, it was observed that the default learning rate (ranging from \emph{1.2e-4} to \emph{1.2e-5}) for GPT-J-6B was inadequate, as illustrated in Fig.\ref{train_loss_1}. To address this, a lower learning rate (ranging from \emph{0.6e-4} to \emph{0.6e-5}) was experimented, and the result is shown by the curve ``6B\_v2''. Subsequent models utilized the same lower learning rate. Additionally, the curve ``6B\_bpe\_25129'' represents the LexGPT-6B model utilizing a domain-specific tokenizer with a vocabulary size of 25129. The curve ``6B\_bpe\_50257'' represents the model using a domain-specific tokenizer with a vocabulary size of 50257.

\begin{figure}[h]
  {\includegraphics[width=0.9\textwidth, keepaspectratio]{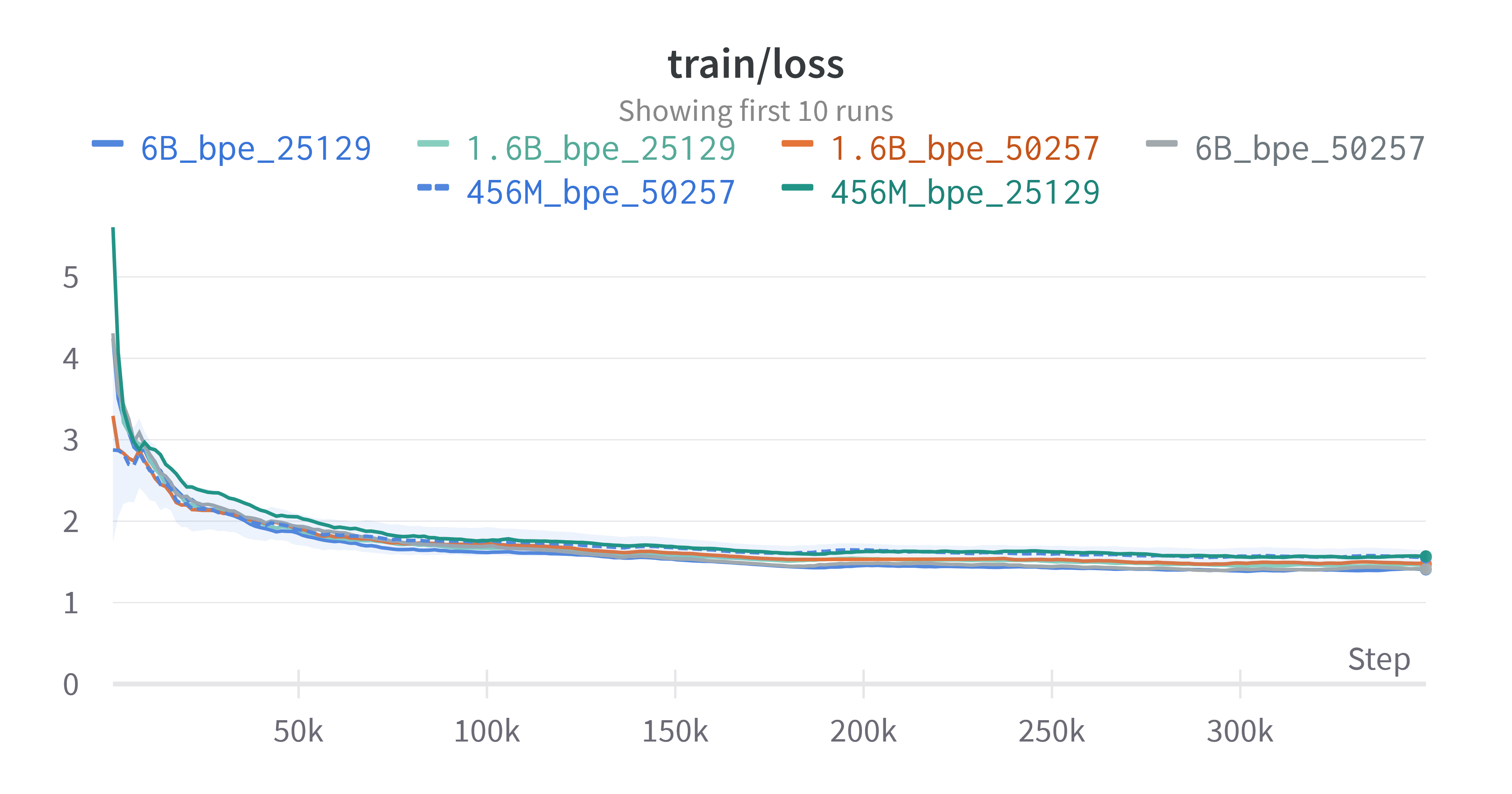}}
  \caption{Training loss}
  \label{train_loss_2}
\end{figure}

Using the domain-specific tokenizers and the same range of lower learning rates, the pre-training process extended to include models of size 1.6B and 456M. The corresponding training losses are displayed in Fig.~\ref{train_loss_2}. Notably, the final training losses did not exhibit significant gains, despite differences in model size. For all models, the training step is 350,000 and the maximum sequence length of the model is 2,048. These settings are the default values in the configuration file ``6B\_roto\_256.json'' provided in GPT-J-6B codebase. In this research, the Pile of Law contains approximately 60 billion tokens after tokenization. By utilizing TPU v3-8 and setting the batch size to 8, the pre-training process covers about 10.6\% of all tokens.

\subsubsection{Release}
Language models specialized in the legal domain have the potential to enhance access to justice. The pre-trained LexGPT models, along with domain-specific tokenizers, tokenized training and validation datasets, configuration files, and relevant source code, will be publicly available upon publication of this manuscript. 
However, it should be noted that language models may make factual mistakes and experience hallucinations. Therefore, legal professionals are recommended as the initial target users of legal applications based on these models, as their legal knowledge can help filter out any mistakes and hallucinations.

\subsection{Downstream Tasks}
\label{subsection:downstream_task}

In this manuscript, the downstream tasks are single-label classification tasks and are achieved by fine-tuning LexGPT models using task-specific text and labels. 
Typically, for classification tasks, additional task-specific layers are added on top of an existing model, and the entire custom setup is fine-tuned end-to-end. This requires modifying the source code of the original model or adding new code to wrap around the extracted body of the model. 
Both involves coding if no classification function is provided in the source code of the original model for customization.
However, one objective of this manuscript is to adhere to the ``No Code'' idea, and thus, the aforementioned coding approaches are not considered.  
Instead, to turn a LexGPT model into a classifier, the model is fine-tuned with training data in the format of ``(text)$<|$label$|>$(label)'', where ``$<|$label$|>$'' is a special tag that concatenates the text and label for training. At inference, the fine-tuned model is utlized to predict the correct label by prompting ``(text)$<|$label$|>$'' to the model. Being able to predict the correct label makes the fine-tuned model a classifier. In this way, one can create a classifier without modifying any source code or structure of the model. 

\subsubsection{Dataset}
The benchmark in this manuscript comes from the LexGLUE~\cite{lexglue}. LexGLUE is based on seven existing legal NLP datasets, selected using criteria largely from SuperGLUE~\cite{SuperGLUE}. The tasks they address have been simplified to make it easier for generic models to address all tasks. 
In this research, the downstream task focuses on the LEDGAR (LEDGAR (Labeled EDGAR) dataset and the CaseHOLD (Case Holdings on Legal Decisions) dataset in LexGLUE. 
As stated in~\cite{lexglue}, the LEDGAR dataset is a dataset for contract provision (paragraph) classification. The contract provisions come from contracts obtained from the US Securities and Exchange Commission (SEC) filings. The original dataset includes approximately 850k contract provisions labeled with 12.5k categories. In LexGLUE, the authors use a subset of the original dataset with 80k contract provisions, considering only the 100 most frequent categories as a simplification. The auhtors split the new dataset chronologically into training (60k, 2016–2017), development (10k, 2018), and test (10k, 2019) sets. Each label represents the single main topic of the corresponding contract provision, i.e., it is a single-label multi-class classification task. The number of classes is 100. As for the CaseHOLD~\cite{case_hold} dataset, it contains approximately 53k multiple choice questions about the holdings of US court cases from the Harvard Law Library case law corpus. The input consists of an excerpt from a court decision, containing a reference to a particular case, where the holding statement is masked out. The task is to identify the correct (masked) holding statement from a selection of five choices. The dataset is split in training (45k), development (3.9k), test (3.9k) sets. 

In this study, the remaining five datasets in LexGLUE are set aside for future experiments due to the following reasons: Firstly, the text in the ECtHR (A and B) and SCOTUS datasets is significantly lengthier than the input sequence length of LexGPT models. In~\cite{lexglue}, the authors employ a hierarchical variant of each pre-trained Transformer-based model that has not been designed for longer text. Since no source code modification is one of the objectives in this manuscript, these three datasets are skipped. Secondly, the EUR-LEX and UNFAIR-ToS datasets are tasks of multi-label classification. Since a generative language model predicts the next token based on previouis tokens, predicting a label may attend inadequately to a previous label in the multi-label settings. Predicting a label should base on its input text only. How to formulate a multi-label classification task and fit the sequential nature of a generative language model is a subject for future research.

\subsubsection{Experiment 1: LEDGAR}

In this experiment, the LexGPT models of size 456M and 1.6B (bpe\_25129) are fine-tuned with the LEDGAR training data in~\cite{lexglue} once. 
The 456M model obtained a \emph{micro-F1} score of 83.5\% and a \emph{macro-F1} score of 72.4\%. 
The 1.6B model obtained a \emph{micro-F1} score of 83.9\% and a \emph{macro-F1} score of 74.0\%.
These numbers are not state-of-the-art results. In~\cite{lexglue}, the highest \emph{micro-F1} score is 88.3\% based on the CaseLaw-BERT model and the highest \emph{macro-F1} score is 83.1\% based on the DeBERTa model. 
It is noted that the original LEDGAR dataset described in~\cite{ledgar} is for multi-label classification. In~\cite{lexglue}, the dataset is simplified as a single-label multi-class classification task. It remains to be investigated in the future whether LexGPT models would outperform the state-of-the-art on the original LEDGAR dataset in a multi-label setting. 

\subsubsection{Experiment 2: CaseHOLD}
In this experiment, the LexGPT models of size 456M and 1.6B (bpe\_25129) are fine-tuned with the CaseHOLD training data once, resulting in accuracies of 49.6\% and 27.6\%, respectively. The CaseHOLD task is to identify the correct holding in a prompt from a selection of five choices. In this study, the multiple choice task is converted into a multiple binary classification task, and the accurarcy is calculated based on the top probability of choices among the multiple binary classification tasks. Although the accuracy (49.6\%) of the 456M model is better than random guesses (20\%), it is still significantly lower than the state-of-the-art result. In~\cite{lexglue}, the CaseHOLD task was performed using the CaseLaw-BERT model, which achieved the highest accuracy of 75.4\%. According to the implementation in~\cite{lexglue}, each training instance consists of the prompt and one of the five candidate answers. The top-level representation h[cls] of each pair is fed to a linear layer to obtain a logit, and the five logits are then passed through a softmax yielding a probability distribution over the five candidate answers. Future research is required to determine whether specialized training data format can help narrow the performance gap of LexGPT models.

\section{Conclusion and Future Work}
\label{section:conclusion}

The major contribution made in this study is the pre-trained LexGPT models using the Pile of Law dataset. The pre-trained LexGPT models will be released at~\cite{github_lexgpt}. Such foundation models can pave the development of InstructGPT-based or ChatGPT-based applications for the legal domain in the future.
In addition, this study aims to provide legal professionals with a simple way to create custom language models without the need to modify its source code. The experimental results demonstrate that the pre-trained LexGPT models can be fine-tuned using task-specific data and labels to function as a classifier. However, despite the minimal effort required, the performance of the fine-tuned GPT model falls short compared to the conventional approach of modifying source code and adding a new classification layer to the model. It is noted that most classification tasks in the legal field are built upon BERT or similar models. How to utilize GPT models are less explored. It remains to be seen whether adding a new classification layer to LexGPT models can outpuform BERT-based models. Another area for future exploration is to investigate, under the ``No Code'' condition, whether the Chain-of-Thought (CoT)~\cite{chainofthought} ability of large language models can enhance the effectiveness of the classifiers in this study if training data is provided in CoT format.

\subsubsection{Acknowledgements} The research reported in this manuscript has been funded by the
Ministry of Science and Technology (MOST) in Taiwan (Project ID:
111-2222-E-A49-005). In addition, the author would like to thank TensorFlow Research Cloud (TRC) greatly for providing powerful computational resources to make this research possible.

\bibliographystyle{splncs04}
\bibliography{citation}

\end{document}